\documentclass{article}[a4]

\usepackage{xurl} 
\usepackage[pdftex]{hyperref}
\usepackage{amsmath}
\usepackage{amssymb}
\usepackage{color}
\usepackage{enumerate}

\begin{document}

%
\title{Designed to Cooperate: A Kant-Inspired Ethic of Machine-to-Machine Cooperation\\
{\color{red} (This preprint has not undergone any post-submission improvements or corrections. \\
The Version of Record of this article is published in {\em AI \& Ethics}, and is available online at \url{https://doi.org/10.1007/s43681-022-00238-5})}}

\author{Seng W. Loke\\School of Information Technology, Deakin University, Geelong, Australia \\
{\tt\small seng.loke@deakin.edu.au}
}
 
\date{}

\maketitle

\begin{abstract}
We envision an increasing presence of devices with agency and autonomous machines in public spaces (e.g., automated vehicles, urban robots and drones) beyond the confines of constrained environments such as a factory floor or research labs. Hence, AI and robotic systems of the future will need to interact with one another, not only in cyber space but also in physical space, and need to behave appropriately in their interactions with one another. 

This commentary highlights an ethic of machine-to-machine cooperation and machine pro-sociality, and argues that  machines capable of autonomous sensing, decision-making and action, such as automated vehicles and urban robots,  owned and used by different self-interested parties, and having their own agendas  (or interests of their owners) should be designed and built to be cooperative in their behaviours, especially if they share public spaces. That is, by design, the machine should first cooperate, and then only consider alternatives if there are problems.

It is argued that being cooperative is not only important for their improved functioning, especially, when they use shared resources (e.g., parking spaces, public roads, curbside space and walkways), but also as a favourable requirement analogous to how humans  cooperating with other humans can be advantageous and often viewed favourably.  
The usefulness of such machine-to-machine cooperation are illustrated via examples including cooperative crowdsourcing, cooperative traffic routing and parking as well as futuristic scenarios involving urban robots for delivery and shopping. 

It is argued that just as  privacy-by-design and security-by-design are important considerations, in order  to yield systems that fulfil ethical requirements, cooperative-by-design should also be an imperative for autonomous systems that are separately owned but co-inhabit the same spaces and use common resources. If a machine using shared public spaces is not cooperative, as one might expect, then it is not only anti-social but not behaving ethically. It is also proposed that certification for urban robots that operate in public could be explored.
 
\end{abstract}


%

\maketitle

\section{Cooperation as an Aspect of Behaving Ethically}

Busy urban spaces can become places of tremendous competition among users. For example, people effectively ``compete'' for parking spaces, or for road use to get from one point to another in the fastest possible way, or with the rise of urban robots, urban robots might compete for space for their operations. For example, urban robots from different companies doing deliveries might compete for space on walkways with other urban robots and people.  Two or more advertising robots might compete for the attention of a  passer-by.
Even robots within the same building might compete with each other for resources while performing tasks or get in the way of each other.

{\color{black}
In this commentary, we mostly use the general meaning of the term ``cooperation'', i.e., ``doing something together'' or ``working together toward a shared aim.''\footnote{From:  \url{https://www.oxfordlearnersdictionaries.com/definition/american_english/cooperation}}  We also consider a Kant-inspired view of cooperation later in this commentary. When a party considers the actions of others in deciding its own action and attempts to apply Kant's categorical imperative (more on this later), we also say that the party has intention to cooperate.
}

However, many (even if not all) such problems can be alleviated by cooperation. For example, road spaces and parking spaces are limited  shared resources in many cities - it was shown that if vehicles could help each other find parking, each vehicle could benefit, or if groups of vehicles cooperate at road junctions in computing their routes, they could all get to their destinations earlier reducing traffic jams~\cite{8818311}. Urban delivery robots could cooperate with each other to ensure safe and efficient use of walkways, instead of trying to out-manoeuvre each other. Vehicles and urban robots could cooperate in sharing information they capture/sense during their journeys, thereby creating a knowledge base for all to benefit from. 

Machine-with-machine cooperation in cyber-space via well-defined protocols is happening everyday in the Internet and the Web with the billions of communicating networked devices.  
Machine-with-machine cooperation (among autonomous machines) in physical space is, however, a consideration only beginning to emerge with the advent of automated vehicles in cities and urban robotics~\cite{SALVINI2018278,doi:10.1177/0042098020917790}.  With an increasing number of (potentially autonomous) machines in public, e.g., automated vehicles and urban robots such as delivery robots on walkways or robots in shopping centres and public spaces, there is then  a growing need for such machines to not only be sensitive to and relate to humans in public spaces, behaving in socially appropriate ways, but also to cooperate and relate to each other in public spaces.

It is noted in~\cite{nagenborg18} that  urban robots occupy physical space, and hence, ``potentially require the rearrangement and redesign of existing spatial entities''. This is significant in cities, especially those with high density, where space is a valuable and scarce resource and there are different land use considerations. It is argued in this commentary that cooperation is an important means of facilitating efficient use of different types of spaces. Robots in cities are not always welcomed, e.g., the delivery robots that use sidewalks in San Francisco were severely restricted in their operations.\footnote{\url{https://www.bbc.com/news/technology-42265048}, though there seems more freedom for deployment in more recent times: \url{https://www.aitrends.com/robotics/last-mile-delivery-robots-making-a-comeback-after-initial-bans/}} Hence, deploying machines  in public spaces can raise issues of space contention, not only with humans but with each other.

Robots that cooperate have been studied in swarm robotics and Internet-based robotics, in particular, where such robots are under the same authority or controller or owner.
But one could envision separately owned autonomous machines that occupy physical spaces in the city and use shared urban resources such as public roads, shared walkways and public spaces,  each with its own (or its owner's) goals, interests and agendas. Regulations can help govern the behaviour of such machines with respect to shared public resources, but, as is  argued here, being aware of each other and cooperating with each other, even helping out each other,  becomes an imperative, not only for their harmonious operation, but also to improve their workings, as we will illustrate further.

This commentary proposes the idea that machines that use shared resources (e.g., roads and public spaces in the city) should be designed to be  cooperative with each other, not only with humans, and that such machines are  behaving ethically when ``first try to cooperate'' is their ``ethos'', by design.

 \section{Lessons from Game Theory}
Towards this direction, one can consider the notion of the {\em Kantian cooperative machine}, based on the concept of Kantian cooperation~\cite{roemer19}, that puts the question of ``us'' as a priority in its decision-making, i.e. if each machine reason in a Kantian manner and ask the same question: ``what strategy(action) would I want all of us to play(take)?'' (call this the {\em K-question}) rather than the question ``what is the best strategy(action) for me to play(take), given others' strategies(actions)?''. 
This notion is not entirely new, considering the Kantian machine discussed in~\cite{10.1109/MIS.2006.77}, where a machine should
``act only according to that maxim whereby you can at the same time will that it should become a universal law'', i.e., ``Kant tells the moral agent to test each maxim (or plan of action) as though it were a candidate for a universalized rule.'' There are examples from game theory, e.g., the Stag Hunt game, that illustrate in theory  the  value of cooperation for achieving Pareto optimality. 

In addition, a default  general rule built into a machine such as ``cooperate when you can and where appropriate'' seems itself to be a good candidate for one such universal rule. 

{\color{black}
The inspiration from Kant is in two ways, yielding two possible principles:
\begin{enumerate}[(i)]
\item  ``cooperate first'' itself is a good candidate for a universalisable maxim (i.e., ``{\em seeking first to cooperate}'' could be willed as {\em a strategy for everyone}), and 
\item intention to cooperate in the sense of each one (individually) asking first the K-question    ``what strategy (action) would I want {\em all} of us to play (take)?''  could be a way to achieve {\em universalisation}, where the ``strategy (action)'' here would depend on the application or domain being considered.
\end{enumerate}
}
 
{\color{black}
Note that the outcomes from the above two could be different.
The argument from game theory must be qualified by consideration of the actual context and settings of the situation. 

For example, consider the payoff matrix (a version of the Stag Hunt game) as follows where the cell corresponding to stag and stag refers to two parties cooperating to catch the stag with payoff $p$ for both parties, the cell corresponding to stag and rabbit (or  rabbit and stag) refers to one party trying to hunt the stag and the other the rabbit, with likely failure for the one trying to hunt the stag alone  (payoff $r$) and likely success for the one catching the rabbit alone (payoff $q$), and the cell for rabbit and rabbit representing the case where each party simply tries to hunt a rabbit on its own, separately.  We have that $p > q > r $ since the payoff $p$ is greatest when both catch the (prized) stag together, and then each working separately to catch a rabbit with payoff  $q$, and then one getting frustrated  trying to catch the stag alone (i.e., $r$) and getting nothing  (not even getting a rabbit).   In this case, cooperation is clearly the best strategy, each getting the highest payoff $p$, and clearly, the strategy is for each party to cooperate.
 \begin{center}
\begin{tabular}{ c|| c| c| }
  & stag (cooperate) & rabbit (independent)  \\ \hline \hline
 stag (cooperate) & p,p & r,q \\   \hline
 rabbit (independent)  & q,r & q,q   \\ \hline
\end{tabular}
\end{center}

In the case above, and in similar cases, where the settings (i.e., the payoffs) are such that cooperation is best, then applying  (i) will result in the best outcomes for both parties, and
using (ii), the two parties  asking the question ``what strategy(action) would I want all of us to play(take)?'' would result in both parties choosing to ``hunt the stag together''  (cooperating), which in this case, is also the most beneficial (since $p > q$).
In this case, both (i) and (ii) leads to the same most beneficial outcome for both parties, which is to cooperate.

However, in other games with different settings, e.g., for suppose it is $p < q$ (somehow the rabbit is bigger and tastier than the stag and can be caught without both parties cooperating!), then applying (i) results in both cooperating (or hunting the stag together) which is a less beneficial outcome for both, but applying (ii) the outcome of each party asking the K-question is to take the ``hunt rabbit individually'' action, which 
still is to uncover the best outcome for both, though the action is not to cooperate! 
Also, we have assumed that the situation is symmetrical, in may be that each party asking the K-question results in different actions because they have their own version of the payoff matrix, but each party is at least trying to apply the categorical imperative.

We have also another qualification which might make it hard to adopt the ``cooperate first'' strategy, even if it results in the best payoffs. This is when the requirements on each party (e.g., by stakeholders) somehow  contradicts the ``cooperate'' action, so that applying (i) is problematic.

In summary, adopting principle (i) clearly predisposes the machines to cooperating, beneficial in certain (qualified) settings as mentioned above. But principle (ii) of asking the K-question might be viewed as intending to cooperate, but which might result in parties later using non-cooperative strategies. 
However, as we will see in the next section, many situations involving such machines cooperating will be beneficial to all, that is, having settings similar to the Stag Hunt game above, especially, when all cooperate rather than only some or none cooperating, and so, principle (i) (and also (ii)) can be applied to the most benefit of all.
}

 \section{Lessons from Urban Applications: Machines Cooperating on Shared Urban Resources}
 One can consider a range of scenarios for machines that are intended to navigate and move autonomously using shared public infrastructure.
 
 \begin{itemize}
\item {\bf Automated Vehicles Cooperating in Routing and Parking.}
 Vehicle-to-vehicle communications (e.g., cooperative-ITS) have been well researched and there are emerging standards (e.g., ETSI\footnote{\url{https://www.etsi.org/technologies/automotive-intelligent-transport}} and SAE\footnote{\url{http://www.sae.org/servlets/pressRoom?OBJECT_TYPE=PressReleases&PAGE=showRelease&RELEASE_ID=3343}}) for formats of messages and protocols for cars to communicate with one another in a variety of contexts, not only to issue warnings for safety, e.g., as a car emerges from a parking space, at intersections or at merging lanes, but also to help vehicles cooperate, e.g., to aid platooning. More recent work is also looking at higher levels of cooperation, instead of simple message exchanges, when finding car parking, or routing to avoid congestion.  Two companies are trialing of  fleets of robotaxis to serve a particular urban areas, though such taxis may not cooperate in the direct  vehicle-to-vehicle communication sense, they are coordinated centrally.\footnote{\url{https://techcrunch.com/2020/08/16/autox-launches-its-robotaxi-service-in-shanghai-competing-with-didis-pilot-program/}}

 \item {\bf Urban Robots Cooperating.}
 Urban robotics have been proposed not only for delivery~\cite{doi:10.1177/0042098020917790}, but also for cleaning streets and waste management. In public retail spaces such as shopping centres, airports, and hotels, there have been deployment of such robots and trials. Police  robots have also been deployed in Singapore.\footnote{E.g., \url{https://www.straitstimes.com/singapore/transport/robot-traffic-cop-spotted-at-changi-airport},\url{https://mothership.sg/2020/05/police-matar-robots/}} There are also security robots playing the role of security guards.\footnote{\url{https://www.knightscope.com}}  The work in~\cite{8794020} proposes the use of swarms of urban robots to collect wastes in the city - such robots can utilise bike lanes. 
  Such robots can cooperate with each other or via a central controller in their tasks.
  
 As such robots occupy valuable urban public real estate, there is a need for their cooperation and coordination - cleaning robots, robots helping to carry shopping, policing robots and delivery robots are, if they proliferate, enough, to cause congestion on sidewalks and public spaces. Also, as mentioned, some types of urban robots are not necessarily welcomed in places predominantly for humans. Delivery robots  have encountered issues in urban environments, as they physically occupy spaces. Similarly,  security robots have been bullied and forced off the street.\footnote{E.g./, see \url{https://www.dezeen.com/2017/12/13/k5-knightscope-security-robot-sfspca-san-francisco-bullied-off-street/}}  Hence, there is a need for coordination and/or cooperation of different types of robots if they are to occupy public spaces traditionally just for humans - perhaps their use of certain areas need to be coordinated or the robots should cooperate to ensure certain constraints on public spaces are satisfied, while performing their functions. 
  
 \item {\bf Machines Contributing to Crowdsourced Knowledge Bases.}
 Crowdsourcing involves aggregating contributions from many to solve problems. In recent times, the idea of crowdsourcing for sensor data has given rise to the concept of crowdsensing, where contributions of mobile sensor data (including geo-tagged pictures, videos, location and other context data) are aggregated to create a overall map or knowledge base of a particular phenomenon. For example, vehicles could collaborate to upload videos to provide a map of where cherry blossoms are in a city~\cite{10.1145/2750858.2804273}.
 
 There are many other uses of crowdsensing, from crowdsourcing in order to create  maps of 4G/5G bandwidth variations to creating maps of where the crowds are in the city, traffic variations, or  temperature~\cite{benmessaoud:tel-01753150,8703108,8566148}. One can imagine machines, whether automated vehicles or urban robots, that roam the streets for their own purposes, but still being able to  provide data towards building real-time urban maps for different purposes, i.e. such machines become contributors towards real-time collective urban knowledge. Such crowdsensing can help inform urban design, e.g., as noted in~\cite{10.1007/978-3-319-69179-440}.
 
\end{itemize}

The examples above motivates cooperation, or at least, coordinated actions among machines.

\section{Lessons from Multiagent Research: Evolving Cooperation and Learning from Each Other}
There is already interesting work on multiagent reinforcement learning where agents learn to optimize not only for their own "self-interest" but the welfare of a group, or to achieve socially optimal outcomes, e.g.,~\cite{zhang19}. Allowing agents to influence each other during learning can lead to better collective  outcomes~\cite{DBLP:conf/icml/JaquesLHGOSLF19}. Social learning is  important to enable agents to learn from each other and to improve themselves, e.g.~\cite{pmlr-v139-ndousse21a,10.1145/2644819}. Even autonomous vehicles could beneficially learn from other autonomous vehicles~\cite{DBLP:journals/corr/abs-2106-05966}.

Machines that use pubic spaces should perhaps have basic ability to learn from each other, and to pick up behavioural ``cues'' from each other, appropriate in a given situation.

\section{Certified Pro-social and Cooperative}
How can we ensure that robots deployed to roam our city streets have been designed and built to be cooperative and pro-social? Is there a way to certify that a given robot will be cooperative, at least within the scope of  its core functions and operational domain? 

For example, when a delivery robot encounters a cleaning robot, they will make way for each other (perhaps according to some pre-programmed manner, and based on some regulatory operational requirements of the local city government), robots from different manufacturers (and belonging to different owners) know how to move in a single file when journeying through  a narrower part of a walkway (even if this means, to an extent, increasing their own travel time) allowing space for pedestrians, robots (separately owned and each with its own agenda) can cooperate in their routing and movement so as not to cause congestion on walkways and public spaces, urban robots might warn each other of hazards encountered, and when a robot breaks down or gets into a dangerous position, other robots come to its aid (which can be in different ways depending on the situation, the type of fault, and the robots' capabilities). 

Even for automated vehicles to exchange context messages, e.g., for safety or platooning, standards are needed and a range of tests required to validate that they conform to particular basic functionalities.
Certification of automated vehicles has been discussed,\footnote{For example, see ~\cite{LEGASHEV2019544}, and  \url{https://unece.org/DAM/trans/doc/2019/wp29grva/GRVA-02-09e.pdf}} and could involve the automated vehicle passing a ``(self-)driving test'' under different operating situations or scenarios. Testing of automated vehicles via simulations has also been considered (e.g., see~\cite{schoener,8835477}).\footnote{See \url{https://www.ansys.com/applications/autonomous-vehicle-validation} and \url{https://www.dlr.de/content/en/articles/news/2021/02/20210430_the-set-level-project.html}}
Testing and validation for pro-sociality and cooperative behaviours for robots could also be carried out, at least to an extent, in a similar way, though it is not clear how complete one can be in terms of the range of   testing situations  - it is likely not possible to be exhaustive, i.e. to anticipate all possible scenarios  that the robots might find itself in.\footnote{A large database of scenarios and situations for automated driving systems testing is being considered: \url{https://www.safetypool.ai}}

\section{\color{black} Conclusion and Future Work}
 As machines interact with humans and other machines in shared public spaces, analogous to how humans live together, machines should be designed to (learn to) ``live'' with each other, and not only humans. Machines can be designed to be cooperative, according to their capabilities, not only to when performing their core functions, but also able to learn from each other (up to what a machine can reasonably be programmed to learn) to behave appropriately in specific contexts. It would be ethical to design such machines to be cooperative, and in being cooperative, as appropriate to the situation, is one aspect of the machines  behaving ethically.
 There could be myriad solutions - for example, there could be a city platform from which robots operating within that city or neighbourhood download ``cooperation rules'' that enable the robots to cooperate in a range of situations. And to be allowed to operate in open shared urban spaces, the robot should somehow first be certified ``pro-social and cooperative''. 
 
 Much future work remains, in order to create the pro-social and cooperative robots  that operate in public spaces shared with other robots and humans, not only the mechanisms, but also the validation of such robots. Work is also needed to determine what would be the basic cooperative requirements on specific types of robots that operate in public. Also, being predisposed to cooperate is not without risks (e.g., even as the Stag Hunt game would illustrate) since the other party might not cooperate (due to fault or malice) and the one trying to cooperate might end up disadvantaged, or might be taken advantage of in the situation.  There is also the issue of the difference between rational cooperation and ethical cooperation -  we have outlined scenarios where it is rational to cooperate, and perhaps from a utilitarian point of view, it would be ethical to cooperate.  In some situations, it might be ethical to cooperate but costly for the individuals to do so.
 
 {\color{black}  Applying principle (i), i.e., seeking first to cooperate, can often be beneficial, especially in settings or situations similar to the {\em Stag Hunt} game settings, but not always.  Determining the actual situations where applying the principle ``seeking first to cooperate'' is beneficial  and where it is not so, would be important, so that the benefits of cooperations can be achieved as often as possible.}
 Future work is required to address these issues.

 Also, future work can perhaps draw lessons for machine-to-machine cooperation by considering work from human-to-human cooperation, e.g., the work on {\em  morality-as-cooperation} might provide a pathway towards developing moral machines; to quote from~\cite{Curry2016}:
 ``...morality as cooperation predicts that people will regard specific types of cooperative behaviour— behaviour that solves some problem of cooperation—as morally good.''
 That is, machines that cooperate can in some way be viewed as moral machines - further thought is required for this.

It is also unclear if the Kantian notion of cooperation here, which asks the question ``what strategy would I want all of us to play?'', can be employed in human-to-machine cooperation, involving a collection of machines and humans. A machine might not have the ability to answer this question adequately. Future work can address this question.  

{\color{black}
\section*{Acknowledgements}
The author would like to thank the anonymous reviewers for the many insightful comments which helped improved the commentary.
}

\bibliographystyle{plain}
\bibliography{refs}

\begin{thebibliography}{10}

\bibitem{8794020}
A.~L. {Alfeo}, E.~C. {Ferrer}, Y.~L. {Carrillo}, A.~{Grignard}, L.~A. {Pastor},
  D.~T. {Sleeper}, M.~G. C.~A. {Cimino}, B.~{Lepri}, G.~{Vaglini}, K.~{Larson},
  M.~{Dorigo}, and A.~{Pentland}.
\newblock Urban swarms: A new approach for autonomous waste management.
\newblock In {\em 2019 International Conference on Robotics and Automation
  (ICRA)}, pages 4233--4240, 2019.

\bibitem{8835477}
Vimal~Rau Aparow, Apratim Choudary, Giridharan Kulandaivelu, Thomas Webster,
  Justin Dauwels, and Niels~de Boer.
\newblock A comprehensive simulation platform for testing autonomous vehicles
  in 3d virtual environment.
\newblock In {\em 2019 IEEE 5th International Conference on Mechatronics System
  and Robots (ICMSR)}, pages 115--119, 2019.

\bibitem{benmessaoud:tel-01753150}
Rim Ben~Messaoud.
\newblock {\em {Towards efficient mobile crowdsensing assignment and uploading
  schemes}}.
\newblock Theses, {Universit{\'e} Paris-Est}, July 2017.

\bibitem{8703108}
A.~{Capponi}, C.~{Fiandrino}, B.~{Kantarci}, L.~{Foschini}, D.~{Kliazovich},
  and P.~{Bouvry}.
\newblock A survey on mobile crowdsensing systems: Challenges, solutions, and
  opportunities.
\newblock {\em IEEE Communications Surveys Tutorials}, 21(3):2419--2465, 2019.

\bibitem{Curry2016}
Oliver~Scott Curry.
\newblock {\em Morality as Cooperation: A Problem-Centred Approach}, pages
  27--51.
\newblock Springer International Publishing, Cham, 2016.

\bibitem{10.1145/2644819}
Jianye Hao, Ho-Fung Leung, and Zhong Ming.
\newblock Multiagent reinforcement social learning toward coordination in
  cooperative multiagent systems.
\newblock {\em ACM Trans. Auton. Adapt. Syst.}, 9(4), December 2014.

\bibitem{DBLP:conf/icml/JaquesLHGOSLF19}
Natasha Jaques, Angeliki Lazaridou, Edward Hughes, {\c{C}}aglar
  G{\"{u}}l{\c{c}}ehre, Pedro~A. Ortega, DJ~Strouse, Joel~Z. Leibo, and Nando
  de~Freitas.
\newblock Social influence as intrinsic motivation for multi-agent deep
  reinforcement learning.
\newblock In Kamalika Chaudhuri and Ruslan Salakhutdinov, editors, {\em
  Proceedings of the 36th International Conference on Machine Learning, {ICML}
  2019, 9-15 June 2019, Long Beach, California, {USA}}, volume~97 of {\em
  Proceedings of Machine Learning Research}, pages 3040--3049. {PMLR}, 2019.

\bibitem{LEGASHEV2019544}
L.V. Legashev, T.V. Letuta, P.N. Polezhaev, A.E. Shukhman, and Yu.A. Ushakov.
\newblock Monitoring, certification and verification of autonomous robots and
  intelligent systems: Technical and legal approaches.
\newblock {\em Procedia Computer Science}, 150:544--551, 2019.

\bibitem{8818311}
Seng~W. Loke.
\newblock Cooperative automated vehicles: A review of opportunities and
  challenges in socially intelligent vehicles beyond networking.
\newblock {\em IEEE Transactions on Intelligent Vehicles}, 4(4):509--518, 2019.

\bibitem{10.1145/2750858.2804273}
Shigeya Morishita, Shogo Maenaka, Daichi Nagata, Morihiko Tamai, Keiichi
  Yasumoto, Toshinobu Fukukura, and Keita Sato.
\newblock Sakurasensor: Quasi-realtime cherry-lined roads detection through
  participatory video sensing by cars.
\newblock In {\em Proceedings of the 2015 ACM International Joint Conference on
  Pervasive and Ubiquitous Computing}, UbiComp '15, page 695–705, New York,
  NY, USA, 2015. Association for Computing Machinery.

\bibitem{nagenborg18}
Michael Nagenborg.
\newblock Urban robotics and responsible urban innovation.
\newblock {\em Ethics and Information Technology}, 2018.

\bibitem{pmlr-v139-ndousse21a}
Kamal~K Ndousse, Douglas Eck, Sergey Levine, and Natasha Jaques.
\newblock Emergent social learning via multi-agent reinforcement learning.
\newblock In Marina Meila and Tong Zhang, editors, {\em Proceedings of the 38th
  International Conference on Machine Learning}, volume 139 of {\em Proceedings
  of Machine Learning Research}, pages 7991--8004. PMLR, 18--24 Jul 2021.

\bibitem{8566148}
J.~{Phuttharak} and S.~W. {Loke}.
\newblock A review of mobile crowdsourcing architectures and challenges: Toward
  crowd-empowered internet-of-things.
\newblock {\em IEEE Access}, 7:304--324, 2019.

\bibitem{10.1109/MIS.2006.77}
Thomas~M. Powers.
\newblock Prospects for a kantian machine.
\newblock {\em IEEE Intelligent Systems}, 21(4):46–51, July 2006.

\bibitem{roemer19}
John~E. Roemer.
\newblock {\em {How We Cooperate: A Theory of Kantian Optimization}}.
\newblock Yale University Press, 2019.

\bibitem{SALVINI2018278}
Pericle Salvini.
\newblock Urban robotics: Towards responsible innovations for our cities.
\newblock {\em Robotics and Autonomous Systems}, 100:278 -- 286, 2018.

\bibitem{schoener}
Hans-Peter Schoener.
\newblock The role of simulation in development and testing of autonomous
  vehicles.
\newblock 09 2017.

\bibitem{doi:10.1177/0042098020917790}
Aidan~H While, Simon Marvin, and Mateja Kovacic.
\newblock {Urban robotic experimentation: San Francisco, Tokyo and Dubai}.
\newblock {\em Urban Studies}, 2020.
\newblock https://doi.org/10.1177

\bibitem{10.1007/978-3-319-69179-440}
Shili Xiang, Lu~Li, Si~Min Lo, and Xiaoli Li.
\newblock People-centric mobile crowdsensing platform for urban design.
\newblock In Gao Cong, Wen-Chih Peng, Wei~Emma Zhang, Chengliang Li, and Aixin
  Sun, editors, {\em Advanced Data Mining and Applications}, pages 569--581,
  Cham, 2017. Springer International Publishing.

\bibitem{zhang19}
Chengwei Zhang, Xiaohong Li, Jianye Hao, Siqi Chen, Karl Tuyls, Wanli Xue, and
  Zhiyong Feng.
\newblock Sa-iga: a multiagent reinforcement learning method towards socially
  optimal outcomes.
\newblock {\em Autonomous Agents and Multi-Agent Systems}, 33(4):403--429,
  2019.

\bibitem{DBLP:journals/corr/abs-2106-05966}
Jimuyang Zhang and Eshed Ohn{-}Bar.
\newblock Learning by watching.
\newblock {\em CoRR}, abs/2106.05966, 2021.

\end{thebibliography}

\end{document}